# Contraction Mapping of Feature Norms for Classifier Learning on the Data with Different Quality

Weihua Liu[1], Xiabi Liu[1*], Murong Wang[1], Ling Ma[2]

[1] Beijing Lab of Intelligent Information Technology, School of Computer, Beijing Institute of Technology, Beijing 100081, China
[2] College of Software, NanKai University, Tianjin 300071, China

**Abstract**

The popular softmax loss and its recent extensions have achieved great success in the deep learning-based image classification. However, the data for training image classifiers usually has different quality. Ignoring such problem, the correct classification of low quality data is hard to be solved. In this paper, we discover the positive correlation between the feature norm of an image and its quality through careful experiments on various applications and various deep neural networks. Based on this finding, we propose a contraction mapping function to compress the range of feature norms of training images according to their quality and embed this contraction mapping function into softmax loss or its extensions to produce novel learning objectives. The experiments on various classification applications, including handwritten digit recognition, lung nodule classification, face verification and face recognition, demonstrate that the proposed approach is promising to effectively deal with the problem of learning on the data with different quality and leads to the significant and stable improvements in the classification accuracy.

## 1 Introduction

In recent years, the performance of classification systems has been significantly improved using deep neural networks trained with softmax loss. The classical softmax loss is good at optimizing the inter-class difference. But it ignores to reduce the intra-class variation. Recently, this shortcoming is improved by adding the margin to the angle between the feature vector and the weight vector [1], or constraining the feature norms to be fixed [2] and normalizing the weights to 1 [3], or utilizing these three strategies simultaneously [4, 5].

Although the above extensions of softmax loss can reduce the intra-class variance and thus improve the overall classification accuracy, they still suffer from a small amount of hard-classified samples (simplified as "hard samples" in the following) like classical softmax loss. This problem cannot be solved by increasing the training iterations. The reason behind such phenomenon is the ignorance to the difference of data in quality in the training or, in other words, the data with different quality are treated equally in the training. This brings disadvantages to the low quality data, since they are harder to be identified than good quality data.

In this paper we discover that there is a positive correlation between the quality of a sample and its feature norm learned with softmax loss. We conduct the experiments on three different applications with three different deep neural networks to reveal this fact, and inspired by which, we propose a contraction mapping function of feature norms and use it to develop novel softmax losses. By using our contraction mapping function, the feature norms of training samples are transformed to a narrower range according to their quality. The good quality data's feature norm will be decreased and the low quality data's feature norm will be increased, under the constraint that the feature norm of low quality data is still less than that of good quality one. Such contraction mapping of feature norms brings the following novel advantages to the learning of classifiers. First, the low quality data will receive more attentions in the training, instead of be treated equally with good quality data. Second, the difference of good and low quality data in learned feature norm will be decreased to reduce the intra-class variance. Third, it is a general learning strategy and can be easily integrated into the loss definitions, for example, integrated into classical and large margin based softmax losses in this paper. Because of these advantages, the resultant new softmax losses can dramatically improve the classification accuracy for low quality data and brings the better results for good quality one. In summary, this paper contributes to the following aspects:

1) To the best of our knowledge, this is the first attempt to consider the problem of data quality difference in the design of learning objective for deep neural networks;

2) The positive correlation between the data's quality and its feature norm learned from the softmax loss are discovered;

3) A new learning idea based on the contraction mapping of feature norms is proposed for dealing with the problem of data quality difference;

4) The proposed contraction mapping of feature norms is embedded into classical and large margin based softmax losses to produce new learning objectives, which lead to significant and stable performance boost in their applications to various classification tasks with various types of deep neural networks.

## 2 Related Works

The work reported in this paper origins from the hard sample problem and proceeds by proposing new softmax losses though considering the correlation between the data's quality and its feature norm. Thus, in this section, we briefly review three related topics: 1) the hard sample mining, 2) the correlation between data quality and feature norm, and 3) the softmax loss and its state-of-the-art extensions.

### 2.1 Hard Sample Mining

We can mine hard samples in the training for improving the generalization performance on borderline cases. Shrivastava et al. [6] selected the samples with top-ranked losses as hard samples and utilized only hard samples in the back-propagation training of neural networks. Xue et al. [7] defined the samples with significant error as the hard samples and employed the enhanced deep multiple instance learning technique to mine them. Sheng et al. [8] proposed a loop network with a ranking list to choose hard samples globally based on the classification difficulty. Suh et al. [9] identified the hard negative classes for an anchor instance based on the class-to-sample distances and then draw hard instances only from the selected classes. For person re-identification, Zhu et al. [10] took the imposter samples as hard ones, by leveraging which a distance metric is learned with the symmetric triplet constraint.

### 2.2 Data Quality & Feature Norm

Parde et al. [11] first revealed through the experiments that the information about the image quality is available in the features embedded in deep convolutional neural network. They concluded that the images in the center of the feature space were uniformly of low quality. Ranjan et al. [2] observed that the norms of face features learned from the softmax loss are informative of the quality of face images. Chen et al. [12] found out that the low-quality iris samples usually have smaller L2-norm values. In this paper, we first explicitly discover the positive correlation between the image's quality and its feature norm learned with the softmax loss.

### 2.3 Softmax Loss and Its Extensions

Since its introduction into deep learning, the softmax loss has played important roles and become the currently most popular learning objective used in applications. Recently, the researchers tried to further improve the performance of the softmax loss through exploring three factors in it: angular margin, feature norms and weight norms.

L-Softmax [1] used an angular margin penalty to enforce the intra-class compactness and the inter-class separability simultaneously. L2-softmax loss [2] and NormFace [3] constrained the feature norm to be a fixed constant. Such strategy of fixed feature norms was then combined with different types of margins to result in CosFace [4] and ArcFace [5]. Liu et al. [13] investigated the large margin softmax loss with different configurations in speaker verification and introduced the ring loss and the minimum hyperspherical energy criterion to further improve the performance. Zhou et al. [14] proposed a double additive margin softmax loss through imposing an additive margin to both the intra-class angular variation and the inter-class angular variation simultaneously. In [15], the authors introduced the adaptive margin for each class to adaptively minimize intra-class variances.

## 3 The Correlation between Data Quality and Feature Norm

Let $m$ be the training batch size, $c$ be the class number, $x_i$ and $y_i$ be the feature vector and the class of the $i$-th sample, respectively, $W_j$ and $b_j$ be the weight vector and the bias corresponding to the $j$-th neuron in the last fully connected layer, respectively, then the classical softmax loss is defined as

$$L = -\frac{1}{m}\sum_{i=1}^{m} \log \frac{e^{W_{y_i}^T x_i + b_{y_i}}}{\sum_{j=1}^{c} e^{W_j^T x_i + b_j}}, \quad (1)$$

where $w^T x = \|w\|\|x\|\cos\theta$ ($\|\cdot\|$ denotes the L$_2$-norm of a vector).

In general, the discrimination between the true class and false classes is subtler for low quality images, compared with good quality ones. And according to Eq. 1 above, in the cases of correct classification, a larger feature norm will result in the larger differences between the classification score for the true class and those for false classes. These two facts plus the ignorance to the difference of data in quality in the training incur that the feature norms will be learned to be larger for good quality data than low quality one.

To validate the analysis above, we observe the correlation among the quality, feature norms and classification accuracies of the data from various applications, including handwritten digit classification on MNIST dataset [16], face classification on CelebA dataset [17] and lung nodule classification on LIDC-IDRI dataset [18]. The neural networks used for these three applications are LeNet-5 [19], VGG-13 [20] and Resnet-34 [21], respectively. After using the softmax loss to train each of the three neural networks on the corresponding training dataset, the feature norms of each sample in test sets are computed. Then we rank the samples in ascending order of feature norms and divide the samples into two subsets according to their ranking. In our experiments, the top 20% of samples are taken as one subset and the other 80% as another subset, where the percentage '20%' is decided by careful observation. By checking and comparing the quality of samples in these two subsets by human observers, we find out that the quality of the images in the subset with the smaller feature norms are obviously and generally lower than that of the images in the subset with the bigger feature norms. Fig. 1 illustrates the examples of such finding. We further compute the average classification accuracy as well as the average feature norm for the images in the two subsets, respectively. The results are listed in Table 1, which confirms the positive correlation between the feature norm, the quality, and the classification accuracy: the low quality data corresponds to the smaller feature norm and the worse classification accuracy in statistical sense. To sum up, we can conclude that the low quality data is hard to be classified and this can be reflected in its feature norm learned from the softmax loss.

It should be noted that the image quality is affected by so many factors. It is very difficult, if not impossible, to measure the degree of image quality with a golden criterion. Wang et al. [22] provided some insights on why image quality assessment is so difficult and advise that the best assessment way is perhaps to look at the images. As shown in the compared samples in Fig. 1, the visual comparison of image quality can be accepted in general.

Table 1. The average feature norms and average classification accuracies for good and low quality data in three applications

| Application | Sample Quality | Sample Number | Accuracy | Feature Norm |
|---|---|---|---|---|
| Handwritten digits | Good | 8,000 | 99.65% | 189.2 |
| | Low | 2,000 | 99.32% | 65.4 |
| | Overall | 10,000 | 99.58% | 164.4 |
| Faces | Good | 162,079 | 99.61% | 25.45 |
| | Low | 40,520 | 98.32% | 13.71 |
| | Overall | 202,599 | 99.35% | 23.10 |
| Lung nodules | Good | 946 | 96.71% | 7.97 |
| | Low | 237 | 84.86% | 3.42 |
| | Overall | 1183 | 94.34% | 7.06 |

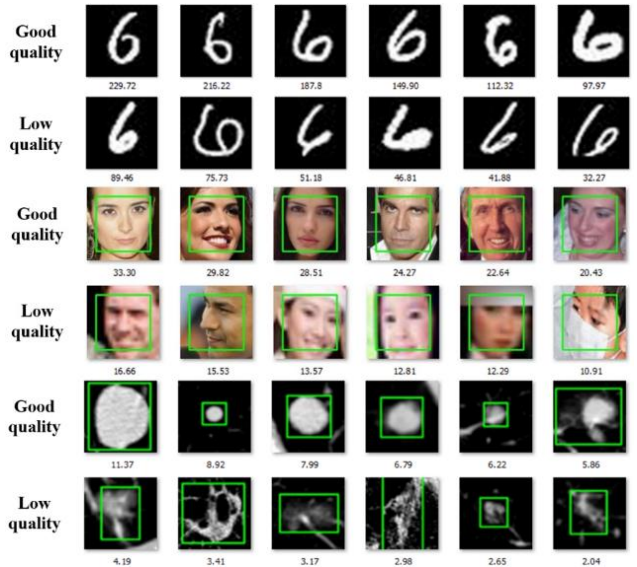

Fig. 1. The visual comparisons of two subsets of images in three applications (the number below each image is its feature norm).

## 4 The Proposed Contraction Mapping and Softmax Losses

According to the observation in the last section, it is possible to increase the feature norms of low quality data for better classification. Under this idea we present our contraction mapping function of feature norms and design the corresponding new softmax losses. They are detailed as follows.

### 4.1 New Softmax Losses

As shown in Eq. 1, $W_j^T x_i + b_j$ is the core for making classification decisions, which is now well-known as target logit [23]. Liu et al. [1] proved that the bias is unnecessary, so we can fix it to 0 [24] and correspondingly simplify the target logit to

$$W_j^T x_i = \|W_j\| \|x_i\| \cos\theta_{W_j, x_i}. \quad (2)$$

Furthermore, the weights can be and are usually normalized to 1, i.e., $\|W_j\| = 1$. Thus, the classification depends on only the feature norm and the angle between the feature vector and the weight vector. Consequently, we have

$$L = -\frac{1}{m} \sum_{i=1}^{m} \log \frac{e^{\|x_i\|\cos(\theta_{W_{y_i}, x_i})}}{e^{\|x_i\|\cos(\theta_{W_{y_i}, x_i})} + \sum_{j=1, j \neq y_i}^{c} e^{\|x_i\|\cos(\theta_{W_j, x_i})}}. \quad (3)$$

Let $P_{y_i}$ be the softmax probability of being the true class $y_i$ for the sample $x_i$. Then the gradient of softmax loss w.r.t. the ground truth label is $1 - P_{y_i}$. And according to Eq. 3, a sample with the larger feature norm will produce the larger $P_{y_i}$. This means that the network will back-propagate the smaller gradient for a sample with the larger feature norm in the learning process based on the softmax loss. So, if we want to let low quality data be trained more sufficiently, we should give it the smaller feature norm than good one. But one the other hand, as analyzed and proved in Section 3, the classification accuracy is positively correlated with the quality of data and the corresponding feature norm learned from the softmax loss. Thus, in order to get the better classification accuracy, we should try to decrease the difference between the finally learned feature norms for good and low quality samples. Now we have two seemingly contradictory learning goals: 1) the less difference between feature norms of good and low quality data and 2) the smaller feature norm for low quality data. We solve this contradiction by introducing a contraction mapping function to compress the original range of feature norms to a narrower one. Fig. 2 illustrates such idea of ours. On one hand, the low feature norms will be increased and the high feature norms will be decreased to reduce the difference between them for the better classification. On the other hand, we keep the ordering relation and the proper difference between the low and high feature norms, in order that the low quality data with smaller feature norms can be paid more attention in the training.

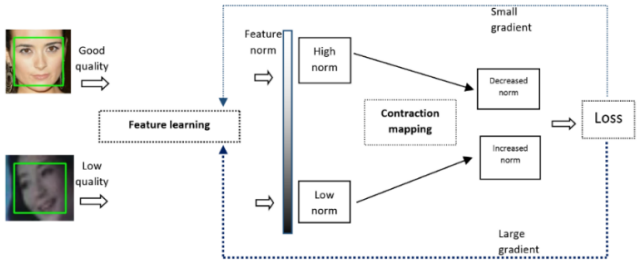

Fig. 2. The illustration of the proposed contraction mapping of feature norms for computing the loss.

The proposed contraction mapping function can be easily integrated into softmax losses to get the better learning objectives. Let $f(\|x\|)$ be the contraction mapping function, the form of which will be explained in the next subsection. The only thing that we need to do is to replace $\|x\|$ with $f(\|x\|)$ in the loss definitions. Taking classical and large margin based softmax losses as two examples, we have

$$L_{CM} = -\frac{1}{m}\sum_{i=1}^{m}\log\frac{e^{f(\|x_i\|)\cos(\theta_{W_{y_i},x_i})}}{e^{f(\|x_i\|)\cos(\theta_{W_{y_i},x_i})}+\sum_{j=1,j\neq y_i}^{c}e^{f(\|x_i\|)\cos(\theta_{W_j,x_i})}}, \quad (4)$$

and

$$L_{CM-M} = -\frac{1}{m}\sum_{i=1}^{m}\log\frac{e^{f(\|x_i\|)\psi(\theta_{W_{y_i},x_i})}}{e^{f(\|x_i\|)\psi(\theta_{W_{y_i},x_i})}+\sum_{j=1,j\neq y_i}^{c}e^{f(\|x_i\|)\cos(\theta_{W_j,x_i})}}, \quad (5)$$

respectively. $\psi(\theta)$ in Eq. 5 can be $\cos(m\theta)$ [24], or $\cos(\theta)-m$ [4], or $\cos(\theta+m)$ [5]. For simplifying the following descriptions, we call our new losses defined in Eq. 4 and Eq. 5 as CM-Softmax and CM-M-Softmax, respectively.

## 4.2 Contraction Mapping

As known from the previous subsection, the contraction mapping function is used to compress the range of feature norms and plays a key role in our new softmax losses. We present a form of the contraction mapping function based on the analysis of the lower and upper bounds of feature norms.

### 4.2.1 Bounds of Feature Norms

For determining the lower and upper bounds of feature norms, we can imagine a constant contraction mapping function that maps all the feature norms to a constant. It actually makes our method degenerate to the fixed-norm based methods [2, 3]. By referring to the lower bound analysis of Ranjan et al. [2] and the upper bound analysis of Yuan et al. [25] for the fixed-norm based softmax losses, we have the following derivation.

1) Lower bound

According to Eq. 3, the average softmax probability of being the true class for correctly classifying a sample is

$$p = \frac{e^{\|x_i\|\cos(\theta_{W_{y_i},x_i})}}{e^{\|x_i\|\cos(\theta_{W_{y_i},x_i})}+\sum_{j=1,j\neq y_i}^{c}e^{\|x_i\|\cos(\theta_{W_j,x_i})}}. \quad (6)$$

Let $s$ be the feature norm of $x_i$, $d$ be the dimension of $x_i$, $c$ be the number of classes.

If $c$ is lower than $2d$, we can distribute the classes on a hypersphere of dimension $d$, such that the weight vectors for any two classes are separated apart by at least $90^0$. Imagine an ideal case that $\theta_{W_{yi},x_i}$ for all the samples and their true classes are $0^0$. Then in this case, the minimum $\theta_{W_{yi},x_i}$ will be $90^0$ for any sample and any one of its false classes. Consequently, only three angles, $0^0$, $90^0$ and $180^0$, can appear in Eq. 6. Fig. 3(a) illustrates such an example for 4 classes in a 2-D feature space. Accordingly, we can transform Eq. 6 to

$$p = \frac{e^s}{e^s+e^{-s}+c-2}. \quad (7)$$

$s$ is generally greater than 1, so $e^{-s}$ will be very small for sufficient high $s$ and can be ignored safely in Eq. 7, then we have $p = e^s/(e^s+c-2)$, according to which we can derive the lowest $s$ for achieving a designated $p$ to be

$$s_{lower} = \log\frac{p(c-2)}{1-p}. \quad (8)$$

If $c$ is larger than $2d$, the angle between the weight vectors for two classes could be less than $90^0$. Thus, we need the larger $s$ to get the same designated value of $p$, compared with the above ideal case that the weight vectors for any two classes are separated apart by at least $90^0$. This means that Eq. 8 gives the lower bound of feature norms for any cases.

2) Upper bound

A theoretical upper bound can be formulated by the lower bound. Fig. 3(b) illustrates the relation between the two bounds, where $s_{upper}$ be the upper bound, $W_1$ and $W_2$ denote the weight vectors for two classes as well as the feature vectors of the corresponding samples (because $\theta_{W_{y_i}, x_i}$ for all the samples and their true classes are $0^0$). Suppose the two classes are separated apart by the angle $\theta$. Then the maximum distance for the features within each class can be represented to be the Euclidean length of the green line segment in Fig. 3: $d_{max-intra} = s_{upper} - s_{lower}$. And the minimum distance for the features between two classes can be represented to be the Euclidean length of the red line segment in Fig. 3: $d_{min-inter} = 2s_{lower}\sin(\theta/2)$. So, to ensure $d_{max-intra} < d_{min-inter}$, we should have $s_{upper} < 2s_{lower}\sin(\theta/2) + s_{lower}$, according to which a reasonable $s_{upper}$ can be set to be $3s_{lower}$.

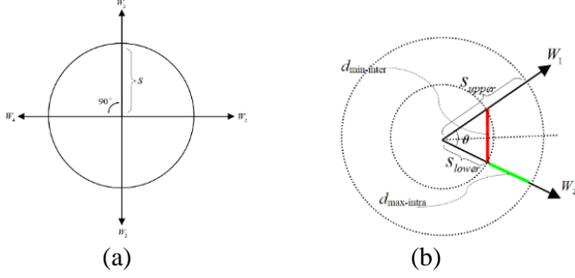

(a)  (b)

Fig. 3. Illustration for the determination of the bounds of feature norms: (a) the lower bound; (b) the upper bound

### 4.2.2 Function Definition

The lower and upper bounds of feature norms provide us a proper range. We will compress the range of original feature norms to such a new range under the constraint that the ordering relation between features norms should be kept, i.e., the new values for original larger feature norms should still be larger. Accordingly, we design the following contraction mapping function to realize the expected compression:

$$f(\|x\|) = s_{lower} + (2 \times sigmoid(\gamma\|x\|) - 1) \times (s_{upper} - s_{lower}), \quad (9)$$

where the logistic sigmoid function is used, $s_{upper}$ and $s_{lower}$ are the bounds analyzed in the last subsection, $\gamma$ is a parameter to control the compression intensity. We should set the larger $\gamma$ for the case with more low quality data and the smaller one for the opposite case.

According to Eq. 9, our contraction mapping function is a monotone increasing function of feature norms and limited in the range $[s_{lower}, s_{upper}]$, which totally satisfies our design goal described above. Furthermore, the growth rate of the function is decreased along with the increasing feature norms. Then in the process of increasing feature norms in the learning, low feature norms will be improved quickly, while the increase of high feature norms will be suppressed sufficiently. This effect is also expected by us.

Our contraction mapping function can be easily implemented and added into the neural networks. Actually, we only need to embed a block between the last two layers, which consists of a normalization layer for obtaining the unit feature vector, a contraction mapping layer for compressing the norm of the feature vector, and a scale layer for multiplying the two results. Fig. 4 shows our implementation in this paper.

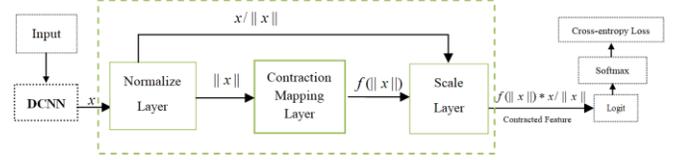

**Figure 4.** The implementation of our contraction mapping function in deep neural networks

### 4.3 Comparison with Fixed Norm

Our method tries to improve the softmax loss by adjusting feature norms, which is similar with the fixed-norm based methods [2, 3]. But we transform the feature norms of all the samples to a narrower range, instead of a single fixed value.

According to the analysis in Section 4.1, if we constrain the feature norms to be fixed for each training sample, the gradients for good and low quality samples will stay at the same level. It will make low quality samples learned insufficiently and thus weaken the learning effect that the methods originally want to achieve. Through applying our contraction mapping function presented in Eq. 9, on one hand, the feature norms will be compressed to reduce the intra-class variation; on the other hand, since the feature norm of low quality data is still lower than that of good quality one, it will be given more gradients during the back-propagation process and thus will be learned more sufficiently than good quality data.

## 5 Experiments

We evaluate the proposed CM-Softmax and CM-M-Softmax losses in four tasks: handwritten digit classification, lung nodule classification, face verification and face recognition. In all the experiments, two parameters in our contraction mapping function are set to be 0.9 ($p$ in Eq. 8) and 1.0 ($\gamma$ in Eq. 9) by experience, respectively. Furthermore, we use the same method described in Section 3 to determine good and low quality data in test sets and measure the performance on the entire set, the good subset and the low subset, respectively.

Our losses are compared with classical softmax loss (simplied as Softmax in the following), NormFace [3] and ArcFace [5]. As described in Section 2, NormFace is a representative of extending Softmax through fixing the feature norms, which can be thought as a counterpart of our CM-Softmax. While ArcFace can be thought as a counterpart of our CM-M-Softmax, since it is a representative of combining the two ideas of large margin separation and fixing norms. The parameters of NormFace and ArcFace, including the fixed feature norm ($s$) and the angular margin ($m$), are set up in each of four tasks, respectively, by careful experiments and referring to the setting in ArcFace.

## 5.1 Handwritten Digit Recognition

The experiments of handwritten digit classification are conducted on famous MNIST dataset [16], which contains 50,000 train examples and 10,000 test examples for 10 digits. LeNet++ [26] is employed as the backbone feature extractor and tailored to output 2-D features for easy visualization of the learned features. We repeat the training and test five times by using each of five losses, respectively. The parameter $s$ in NormFace and ArcFace is 10 and $m$ in ArcFace is 0.5.

The resultant 2-D features of 10,000 test samples in one test are depicted in Fig. 5, where each of the lobes corresponds to a digit class. Comparing Fig. 5(b) with Fig. 5(c) and comparing Fig. 5(d) with Fig. 5(e), we can see that introducing the contraction mapping of feature norms brings not only much better intra-variation but also obvious reduction of the features with tiny norms.

Table 2 lists the means of average classification accuracies over five tests. As shown in Table 2, CM-Softmax surpasses NormFace thoroughly and CM-M-Softmax surpasses ArcFace thoroughly. Especially on the low quality data, the improvement is more significant. According to these results and the comparisons in Fig. 5, we can conclude that the proposed contraction mapping of feature norms is valuable for effectively improving the learning on low quality data without the sacrifice of the performance on good quality data.

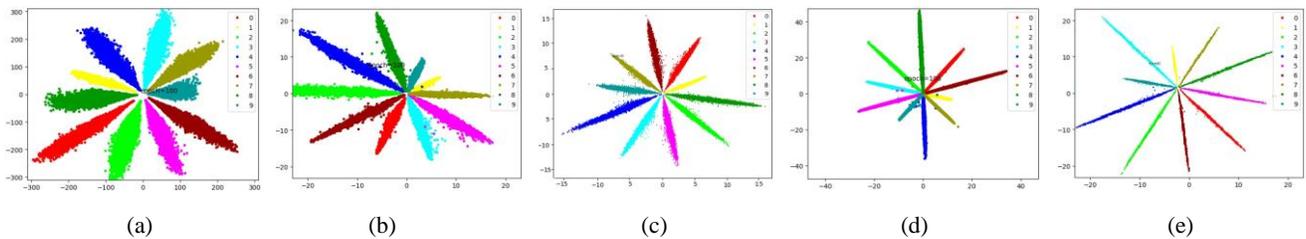

(a) (b) (c) (d) (e)

Figure 5. Visualization of 2-D features of the test samples in MNIST, learned by using (a) Softmax; (b) NormFace; (c) CM-Softmax; (d) ArcFace; (e) CM-M-Softmax

Table 2. Classification Accuracies on MNIST test set

| Data | Softmax | NormFace | CM-Softmax | ArcFace | CM-M-Softmax |
|---|---|---|---|---|---|
| Good Quality | 98.96% | 99.16% | 99.18% | 99.19% | **99.21%** |
| Low Quality | 98.27% | 98.49% | 99.01% | 98.87% | **99.12%** |
| Entire | 98.82% | 99.03% | 99.15% | 99.13% | **99.19%** |

## 5.2 Lung Nodule Classification

The LIDC-IDRI dataset [18] is a widely used evaluation dataset for lung nodule detection and classification. In LUNA 16 challenge [27], 641,822 lung nodule candidates extracted from the CT scans in LIDC-IDRI are provided. The classification task is to detect the true nodules from these 641,822 candidates. We realize the classification by using each of five losses to train multi-level contextual 3D ConvNets [28], respectively. The parameter $s$ in NormFace and ArcFace is 5 and $m$ in ArcFace is 0.2.

The resultant performance is measured by Competition Performance Metric (CPM) introduced in the ANODE09 Challenge [29]. Firstly, the detection sensitivity (the number of identified true nodules divided by the number of ground truth nodules) and the average false positive rate per scan (FPs/scan) are recorded. Then the CPM is calculated as the average of the sensitivities at seven predefined FPs/scan (1/8, 1/4, 1/2, 1, 2, 4, and 8). The Free-Response Operating Characteristic (FROC) curve [30] can reflect the change of sensitivities as a function of FPs/scan. Fig. 6 shows the FROC curves from five losses. The corresponding CPMs and the sensitivities under 1FP/scan (Sen/1FP) are listed in Table 3. Fig. 6 and Table 3 shows that CM-M-Softmax behaves the best and surpasses ArcFace with the increase rate of 0.74% in CPM, and CM-Softmax surpasses NormFace with the increase rate of 2.63% in CPM. It again proves the value of our contraction mapping function.

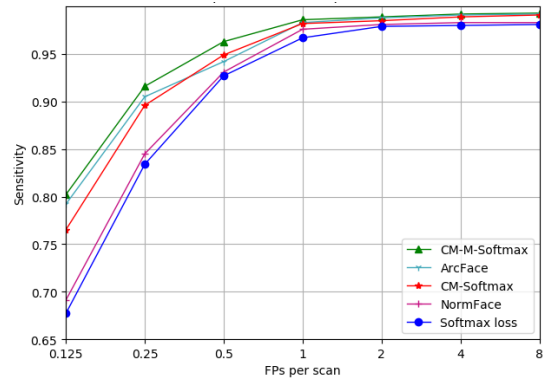

Figure 6. FROC curves on the LIDC-IDRI dataset for three losses

Table 3. The performance of five losses for lung nodule classification

| Loss | Sen/1FP | CPM | Increase Rate in CPM |
|---|---|---|---|
| Softmax loss | 0.967 | 0.906 | 4.47% |
| NormFace | 0.976 | 0.913 | 3.94% |
| CM-Softmax | 0.982 | 0.937 | 1.28% |
| ArcFace | 0.983 | 0.942 | 0.74% |
| **CM-M-Softmax** | **0.986** | **0.949** | / |

## 5.3 Face Verification

In this application, a training dataset that contains 4.2M images of 87k unique identities is constructed from MS-Celeb-1M [31] by using the method adopted in ArcFace. As for the test datasets, we use 1) IJB-B [32] that contains 1,845 subjects with 10,270 genuine matches and 8M impostor matches and 2) 5,000 video pairs in YouTube Face (YTF) [33], following the unrestricted, labeled outside data protocol [34] . Furthermore, we follow the method in SphereFace [24] to generate the normalized face crops (128×128) by utilizing five facial points, for all the training and test samples.

The ResNet-50 network [21] is used as our backbone feature extractor. And during the set-to-set testing, to get the features for videos (YTF) or templates (IJB-B), we calculate the feature mean of all images from the same identity. The parameter $s$ in NormFace and ArcFace is 32 and $m$ in ArcFace is 0.5.

Table 4 lists the face verification accuracies resulted from the five losses. Accordingly, we have the following observations: 1) CM-M-Softmax behaves the best on all the tests; 2) CM-Softmax behaves better thoroughly than NormFace; 3) the performance on low quality data is improved by a large margin. Compared with ArcFace, the increase rates of 12.16% and 22.86% in error rate are brought by CM-M-Softmax in two test sets, respectively. Compared with NormFace, the increase rates of 35.43% and 33.33% in error rate are brought by CM-Softmax in two test sets, respectively.

Table 4. Face verification performance (%) from five losses on YTF and IJB-B datasets

| Loss | Data Set | YTF | IJB-B |
|---|---|---|---|
| Softmax | Good Qaulity | 97.6 | 97.4 |
| | Low Qaulity | 84.7 | 86.1 |
| | Entire | 93.7 | 94.8 |
| NormFace | Good | 98.7 | 98.9 |
| | Low | 87.3 | 89.8 |
| | Entire | 96.3 | 97.3 |
| CM-Softmax | Good | 99.1 | 99.2 |
| | Low | 91.8 | 93.2 |
| | Entire | 97.2 | 98.1 |
| ArcFace | Good | 99.2 | 99.3 |
| | Low | 92.6 | 93.0 |
| | Entire | 97.9 | 98.0 |
| CM-M-Softmax | Good | **99.3** | **99.5** |
| | Low | **93.5** | **94.6** |
| | Entire | **98.1** | **98.5** |

## 5.4 Face Recognition

For conducting face recognition experiments, we use the same training dataset as that in Section 5.3 and select famous MegaFace challenge dataset [31] as the test set. It includes a gallery set and a probe set. The gallery set contains more than one million images from 690k individuals. The probe set consists of FaceScrub dataset (100k photos of 530 unique individuals) and FGNet dataset (1,002 face ageing images from 82 identities). We remove the overlapped face images from one million distractors in MegaFace, as done in ArcFace. MegaFace challenge has two protocols including large or small training sets. We choose the large protocol in the tests.

For the backbone feature extractor, we employ a recent variant of ResNet-101 architecture that adopted in ArcFace. The parameter $s$ in NormFace and ArcFace is 64 and $m$ in ArcFace is 0.5, following the setting in ArcFace.

We give the recognition results in Table 5, where "Rank 1@10$^6$" denotes the rank-1 face identification accuracy with 1M distractors and "TAR@FAR10$^{-6}$" denotes the true accepted rate at 10$^{-6}$ false accepted rate. We can see that CM-Softmax and CM-M-Softmax obviously behaves better than their counterparts, with relative performance gains of 66.37% and 48.55% in rank-1 error rate, respectively. In summary, our contraction mapping function of feature norms leads to significant and stable improvements in all the four tasks.

Table 5. Face recognition results (%) of five losses on MegaFace challenge

| Index | Softmax | NormFace | CM-Softmax | ArcFace | CM-M-Softmax |
|---|---|---|---|---|---|
| Rank1@10$^6$ | 91.84 | 94.32 | 98.09 | 98.21 | **99.06** |
| VR@FAR10$^{-6}$ | 94.38 | 97.29 | 98.14 | 98.27 | **99.11** |

## 6 Conclusions

In this paper, we have revealed the positive correlation between the data's quality and its feature norm learned from the softmax loss and correspondingly, proposed a contraction mapping function to compress the range of feature norms and embedded it into the softmax loss or its extensions to produce novel learning objectives. The theoretical analysis and the experiment results on various applications and various deep neural networks demonstrate that introducing our contraction mapping function of feature norms are promising to effectively deal with the problem of the difference of data in quality and thus can bring significant and stable performance boost to the learning based on the softmax losses.

# References

1. Liu W, Wen Y, Yu Z, Yang M: **Large-Margin Softmax Loss for Convolutional Neural Networks**. In: *ICML: 2016*. 507-516.
2. Ranjan R, Castillo CD, Chellappa R: **L2-constrained softmax loss for discriminative face verification**. *arXiv preprint arXiv:170309507* 2017.
3. Wang F, Xiang X, Cheng J, Yuille AL: **Normface: l 2 hypersphere embedding for face verification**. In: *Proceedings of the 2017 ACM on Multimedia Conference: 2017*. ACM: 1041-1049.
4. Wang F, Cheng J, Liu W, Liu H: **Additive margin softmax for face verification**. *IEEE Signal Processing Letters* 2018, **25**(7):926-930.
5. Deng J, Guo J, Zafeiriou S: **Arcface: Additive angular margin loss for deep face recognition**. *arXiv preprint arXiv:180107698* 2018.
6. Shrivastava A, Gupta A, Girshick R: **Training region-based object detectors with online hard example mining**. In: *Proceedings of the IEEE Conference on Computer Vision and Pattern Recognition: 2016*. 761-769.
7. Xue J, Han J, Zheng T, Guo J, Wu B: **Hard Sample Mining for the Improved Retraining of Automatic Speech Recognition**. *arXiv preprint arXiv:190408031* 2019.
8. Sheng H, Zheng Y, Ke W, Yu D, Cheng X, Lyu W, Xiong Z: **Mining hard samples globally and efficiently for person re-identification**. *IEEE Internet of Things Journal* 2020.
9. Suh Y, Han B, Kim W, Lee KM: **Stochastic class-based hard example mining for deep metric learning**. In: *Proceedings of the IEEE Conference on Computer Vision and Pattern Recognition: 2019*. 7251-7259.
10. Zhu X, Jing X-Y, Zhang F, Zhang X, You X, Cui X: **Distance learning by mining hard and easy negative samples for person re-identification**. *Pattern Recognition* 2019, **95**:211-222.
11. Parde CJ, Castillo C, Hill MQ, Colon YI, Sankaranarayanan S, Chen J-C, O'Toole AJ: **Deep convolutional neural network features and the original image**. *arXiv preprint arXiv:161101751* 2016.
12. Chen Y, Wu C, Wang Y: **T-Center: A Novel Feature Extraction Approach Towards Large-Scale Iris Recognition**. *IEEE Access* 2020, **8**:32365-32375.
13. Liu Y, He L, Liu J: **Large margin softmax loss for speaker verification**. *arXiv preprint arXiv:190403479* 2019.
14. Zhou S, Chen C, Han G, Hou X: **Double Additive Margin Softmax Loss for Face Recognition**. *Applied Sciences* 2020, **10**(1):60.
15. Liu H, Zhu X, Lei Z, Li SZ: **Adaptiveface: Adaptive margin and sampling for face recognition**. In: *Proceedings of the IEEE Conference on Computer Vision and Pattern Recognition: 2019*. 11947-11956.
16. LeCun Y: **The MNIST database of handwritten digits**. *http://yann lecun com/exdb/mnist/* 1998.
17. Liu Z, Luo P, Wang X, Tang X: **Deep learning face attributes in the wild**. In: *Proceedings of the IEEE International Conference on Computer Vision: 2015*. 3730-3738.
18. Armato III SG, McLennan G, Bidaut L, McNitt‐Gray MF, Meyer CR, Reeves AP, Zhao B, Aberle DR, Henschke CI, Hoffman EA: **The lung image database consortium (LIDC) and image database resource initiative (IDRI): a completed reference database of lung nodules on CT scans**. *Medical physics* 2011, **38**(2):915-931.
19. LeCun Y, Bottou L, Bengio Y, Haffner P: **Gradient-based learning applied to document recognition**. *Proceedings of the IEEE* 1998, **86**(11):2278-2324.
20. Simonyan K, Zisserman A: **Very deep convolutional networks for large-scale image recognition**. *arXiv preprint arXiv:14091556* 2014.
21. He K, Zhang X, Ren S, Sun J: **Deep residual learning for image recognition**. In: *Proceedings of the IEEE conference on computer vision and pattern recognition: 2016*. 770-778.
22. Wang Z, Bovik AC, Lu L: **Why is image quality assessment so difficult?** In: *2002 IEEE International Conference on Acoustics, Speech, and Signal Processing: 2002*. IEEE: IV-3313-IV-3316.
23. Hinton G, Vinyals O, Dean J: **Distilling the knowledge in a neural network**. *arXiv preprint arXiv:150302531* 2015.
24. Liu W, Wen Y, Yu Z, Li M, Raj B, Song L: **Sphereface: Deep hypersphere embedding for face recognition**. In: *The IEEE Conference on Computer Vision and Pattern Recognition (CVPR): 2017*. 1.
25. Yuan Y, Yang K, Zhang C: **Feature Incay for Representation Regularization**. *arXiv preprint arXiv:170510284* 2017.
26. Wen Y, Zhang K, Li Z, Qiao Y: **A discriminative feature learning approach for deep face recognition**. In: *European Conference on Computer Vision: 2016*. Springer: 499-515.
27. LUNA16: **Lung nodule analysis 2016**. *https://luna16grand-challengeorg/*.
28. Dou Q, Chen H, Yu L, Qin J, Heng P-A: **Multilevel contextual 3-d cnns for false positive reduction in pulmonary nodule detection**. *IEEE Transactions on Biomedical Engineering* 2017, **64**(7):1558-1567.
29. Niemeijer M, Loog M, Abramoff MD, Viergever MA, Prokop M, van Ginneken B: **On combining computer-aided detection systems**. *IEEE Transactions on Medical Imaging* 2011, **30**(2):215-223.
30. Kundel H, Berbaum K, Dorfman D, Gur D, Metz C, Swensson R: **Receiver operating characteristic analysis in medical imaging**. *ICRU Report* 2008, **79**(8):1.
31. Guo Y, Zhang L, Hu Y, He X, Gao J: **Ms-celeb-1m: A dataset and benchmark for large-scale face

**recognition**. In: *European Conference on Computer Vision: 2016*. Springer: 87-102.
32. Whitelam C, Taborsky E, Blanton A, Maze B, Adams JC, Miller T, Kalka ND, Jain AK, Duncan JA, Allen K: **IARPA Janus Benchmark-B Face Dataset**. In: *CVPR Workshops: 2017*. 6.
33. Wolf L, Hassner T, Maoz I: **Face recognition in unconstrained videos with matched background similarity**. In: *Computer Vision and Pattern Recognition (CVPR), 2011 IEEE Conference on: 2011*. IEEE: 529-534.
34. Huang GB, Learned-Miller E: **Labeled faces in the wild: Updates and new reporting procedures**. *Dept Comput Sci, Univ Massachusetts Amherst, Amherst, MA, USA, Tech Rep* 2014:14-003.